\documentclass[10pt,twocolumn,letterpaper]{article}

\usepackage{iccv}
\usepackage{times}
\usepackage{epsfig}
\usepackage{graphicx}
\usepackage{amsmath}
\usepackage{amssymb}

\usepackage{bm}
\usepackage{booktabs}
\usepackage{multirow}
\usepackage{array}

\usepackage[pagebackref=true,breaklinks=true,letterpaper=true,colorlinks,bookmarks=true]{hyperref}

\iccvfinalcopy 

\def\iccvPaperID{****} 

\begin{document}

\title{UnitModule: A Lightweight Joint Image Enhancement Module for Underwater Object Detection}

\author{
Zhuoyan Liu \quad Bo Wang\thanks{Corresponding author.}
\quad Ye Li \quad Jiaxian He \quad Yunfeng Li\\
Harbin Engineering University\\
{\tt\small \{liuzhuoyan,wb,liye,hjx666,liyunfeng\}@hrbeu.edu.cn}}


\maketitle

\begin{abstract}
Underwater object detection faces the problem of underwater image degradation, which affects the performance of the detector. Underwater object detection methods based on noise reduction and image enhancement usually do not provide images preferred by the detector or require additional datasets. In this paper, we propose a plug-and-play \textbf{U}nderwater joi\textbf{n}t \textbf{i}mage enhancemen\textbf{t} \textbf{Module} (UnitModule) that provides the input image preferred by the detector. We design an unsupervised learning loss for the joint training of UnitModule with the detector without additional datasets to improve the interaction between UnitModule and the detector. Furthermore, a color cast predictor with the assisting color cast loss and a data augmentation called Underwater Color Random Transfer (UCRT) are designed to improve the performance of UnitModule on underwater images with different color casts. Extensive experiments are conducted on DUO for different object detection models, where UnitModule achieves the highest performance improvement of 2.6 AP for YOLOv5-S and gains the improvement of 3.3 AP on the brand-new test set (\(\text{URPC}_{test}\)). And UnitModule significantly improves the performance of all object detection models we test, especially for models with a small number of parameters. In addition, UnitModule with a small number of parameters of 31K has little effect on the inference speed of the original object detection model. Our quantitative and visual analysis also demonstrates the effectiveness of UnitModule in enhancing the input image and improving the perception ability of the detector for object features. The code is available at \url{https://github.com/LEFTeyex/UnitModule}.
\end{abstract}

\begin{figure}[t]
	\centering
		\includegraphics[width=\linewidth]{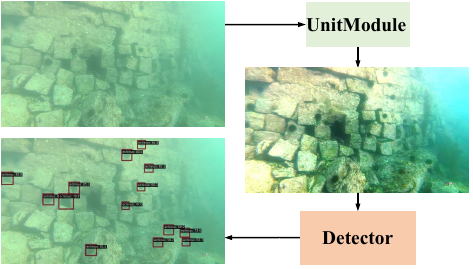}
	\caption{
    The inference flow for the detector with UnitModule. UnitModule is jointly trained with the detector. The enhanced image is only visualized for display, it is actually an intermediate tensor in forward propagation.
    }
	\label{fig:overall}
\end{figure}

\section{Introduction}
\label{sec:Introduction}

Underwater object detection faces significant challenges. Due to the absorption of light of different wavelengths in the water medium and the suspended particles in water, underwater images usually suffer from degradation such as color cast~\cite{3}, blurring, \textit{etc}. We argue that such degradation introduces noise to the image, making it difficult for the object detection network to learn the original features of the object. Some works do not consider the effect of noise on underwater object detection~\cite{4}. The degradation leads to poor performance and the generalization of the detector on different underwater datasets or underwater environments~\cite{14}.

Underwater object detection algorithms are usually deployed on embedded devices where processing power is limited and real-time processing is required. Therefore, lightweight detection models are required underwater. Research shows that the generalization performance of lightweight models is limited~\cite{78}. This limitation makes it difficult for lightweight models to learn about different underwater noise. We believe that the underwater lightweight detection model devotes some of its attention to generalizing noise, which reduces the performance of the model on detection. In this work, we focus on the way of noise reduction that improves the attention of the model to the detection.

For the mentioned problems, some works provide enhanced images for the detector by preprocessing input images using image enhancement methods. They assume that enhanced images improve the performance of the detector in the underwater environment. For example, both~\cite{15} and~\cite{17} use the traditional image enhancement method to enhance images. While ULO~\cite{18} designs a module to predict the hyper-parameters of the traditional image enhancement method for each image. Besides, HybridDetectionGAN~\cite{19} jointly trains the detector with GAN networks to reduce noise for input images. However, these methods might suffer from two limitations:

\textbf{(1) The enhanced images are usually not what the object detector prefers.}
In some underwater environments, preprocessing input images by the traditional image enhancement methods might reduce the performance of the detector because these traditional methods are proposed to improve the human visual perception of images. They ignore the potential information that facilitates object detection and sometimes even generate distorted images that mislead the detector. The image enhancement methods~\cite{19} based on deep learning are able to learn the potential information by training with the detector jointly. Although the interaction between the image enhancement module and the detector is increased, the additional paired image enhancement dataset that is required for assisting training introduces bias into the enhancement network. In other words, the enhanced image is not what the object detector prefers in practical applications.

\textbf{(2) Additional datasets are required.}
The image enhancement methods~\cite{19} that are jointly trained with the detector usually require additional paired image enhancement datasets to support the training of the enhancement module. However, some object detection tasks do not have access to these additional datasets to use these methods.

To address these limitations, we propose an \textbf{U}nderwater joi\textbf{n}t \textbf{i}mage enhancemen\textbf{t} \textbf{Module} (UnitModule), which is a plug-and-play lightweight module to enhance the input image as in Figure~\ref{fig:overall}. UnitModule estimates the transmission map and calculates the global background light to clean up the noise in the input image, which is based on the modified Koschmieder's model~\cite{23}. We design the unsupervised learning loss for UnitModule inspired by~\cite{24} and train the detector with UnitModule jointly without any additional dataset. This Koschmieder's model as a weak constraint can make it easy for UnitModule to learn in the direction that the detector prefers (converge in the direction of detector losses). The main contributions of this paper are as follows:
\begin{itemize}
    \item We propose a plug-and-play lightweight module UnitModule to enhance the input image and design an unsupervised learning loss for its joint training with the detector.

    \item We design a color cast predictor with the assisting color cast loss and a data augmentation called Underwater Color Random Transfer (UCRT) to improve the generalization performance of UnitModule on underwater images with different color casts.

    \item Our extensive experiments on DUO~\cite{25} and \(\text{URPC}_{test}\)\footnote{Underwater Robot Professional Contest (URPC). The datasets URPC2020 and URPC2021 are provided by URPC at \url{https://openi.pcl.ac.cn/OpenOrcinus_orca}}. (details in Section~\ref{sec:Experiments}) validate the effectiveness of our UnitModule and demonstrate that our UnitModule improves the generalization performance of the detector on underwater images.

    \item Our quantitative and visual analysis confirms that our UnitModule improves the image quality and makes the detector focus more on the object at the feature level, reducing the impact of background noise. It also indicates that the image enhanced by UnitModule is what the detector prefers.
\end{itemize}

The rest of this paper is organized as follows: Section~\ref{sec:Related Work} reviews some related works. Section~\ref{sec:Method} describes the details of the proposed method. Section~\ref{sec:Experiments} contains relevant experiments and analysis. Section~\ref{sec:Conclusion} is about conclusion. Section~\ref{appendix:appendix A} contains additional experimental data and results.
\section{Related Work}
\label{sec:Related Work}

\textbf{Object detection.}
In modern times, object detectors based on deep learning are mainly categorized as two-stage detectors and one-stage detectors. Two-stage detectors mainly include R-CNN series~\cite{28}, SPPNet~\cite{30}, \textit{etc}. The most representative one-stage detectors include YOLOv5-8~\cite{36,37,81,82}, YOLOX~\cite{38}, RTMDet~\cite{83}, RetinaNet~\cite{40}, FCOS~\cite{41}, TOOD~\cite{42}, \textit{etc}. In recent years, a new fully end-to-end detector, DETR~\cite{43}, DINO~\cite{84}, has emerged, which removes the non-maximum suppression (NMS) from the above detectors. The above detection methods are also used in underwater object detection. However, the problem of detection in underwater environments where there is noise interference is not taken into account by these methods.

\textbf{Underwater object detection.}
Due to the noise in complex underwater environments, the application of object detection algorithms in underwater scenes is difficult. Some traditional image enhancement methods~\cite{15, 17} are used in object detection to preprocess the input image. The follow-up ULO~\cite{18} designs a module for the dynamic prediction of hyper-parameters in the image enhancement method for each image. These methods are limited by this fixed image enhancement paradigm, which reduces the interaction with the detector and affects the applicability of the detection model. Several similar works use image enhancement methods based on deep learning such as GAN~\cite{19}, CNN~\cite{87}, \textit{etc}. and improve the interaction between detectors and them. GCC-Net~\cite{86} designs a dual-branch backbone to train the underwater object detector using enhanced and raw images as inputs. However, they require additional paired image enhancement datasets, which makes their training difficult. The above methods based on noise reduction do not answer why preprocessing the input image with image enhancement improves the performance of underwater object detection, nor do they indicate the scope of application of these methods. Another kind of method mainly focuses on the generalization performance of the object detection model in the condition that the input image contains different noise. RoIMix~\cite{4} designs a data augmentation to simulate overlapping, occluded, and blurred objects. DMCL~\cite{77} optimizes the training method and applies contrastive learning which is designed to improve the domain generalization performance of the detector. FERNet~\cite{7} proposes a receptive field enhancement module for the backbone to exploit multi-scale semantic features. SWIPENET~\cite{44} designs a sample-weighted hyper network and a robust training paradigm to learn potential information from different noise. Though these works show strong performance, they do not consider the relationship between the object detection model scale and the generalization performance.

\textbf{Modified Koschmieder's model for underwater image enhancement.}
Modified Koschmieder's model~\cite{23} well describes image degradation caused by light scattering and absorption in underwater environments. Some non-deep learning image enhancement methods~\cite{48} based on the modified Koschmieder's model achieve great performance. Besides, there are also image enhancement methods~\cite{52} based on CNN. These image enhancement models inspire the current work to develop an unsupervised image enhancement module that is jointly trained with the detector.
\section{Method}
\label{sec:Method}

\subsection{Overall Architecture of UnitModule}
\label{sec:Overall Architecture of UnitModule}

The UnitModule is based on the inverse process of the modified Koschmieder's model to enhance the image. The modified Koschmieder's model explains the underwater image formation~\cite{23} as follows:
\begin{equation}
\label{eq:km}
    \bm{I}(x) = \bm{J}(x)\bm{t}(x) + \left(1 - \bm{t}(x) \right)\bm{A}
\end{equation}
where \(x\) represents the image pixel, \(\bm{I}\) is the degraded image, \(\bm{J}\) is the undegraded image, \(\bm{A}\) is the global background light underwater which represents the atmospheric light value without particle attenuation (\(\bm{A}\) is called the atmospheric light in the atmosphere). \(\bm{t}\) is the transmission map, in which the physical significance is the proportion of light that reaches the visual sensor after particle attenuation. \(\bm{t}\) is different in the color channels of underwater images~\cite{23}.\newline Transforming Eq.~\eqref{eq:km}, we get:
\begin{equation}
\label{eq:transform km}
    \bm{J}(x) = \left( \bm{I}(x) - (1 - \bm{t}(x))\bm{A} \right) / \bm{t}(x)
\end{equation}
\(\bm{I}\) is the degraded image in the dataset, \(\bm{t}\) and \(\bm{A}\) is estimated by UnitModule. Then we use the formula Eq.~\eqref{eq:transform km} to calculate the enhanced image \(\bm{J}\) which is the input to the detector.

\begin{figure}[t]
	\centering
		\includegraphics[width=\linewidth]{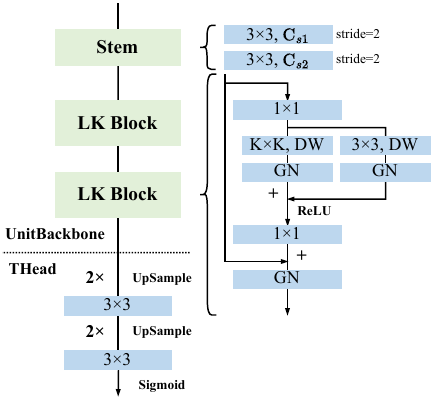}
	\caption{
    The architecture of UnitModule. AHead has no parameter, so it is not displayed. The main architecture includes depth-wise (DW) large kernel convolution (conv), 3\(\times\)3 conv, 1\(\times\)1 conv, and Group Normalization (GN) which the number of groups is 8. The last conv in THead uses the sigmoid activation function but no GN. All other convs have a following GN, some of which are not shown here. We use ReLU after conv-GN sequences, except for those before the addition in the large kernel and those before the shortcut addition.
    }
	\label{fig:architecture}
\end{figure}

An overview of the UnitModule architecture is presented in Figure~\ref{fig:architecture}, which consists of UnitModule Backbone (UnitBackbone), Transmission Head (THead), and Atmosphere Head (AHead) that is not shown. The UnitModule, as a plug-and-play module, is designed to be lightweight to reduce the impact on the inference speed of the detection model. And UnitModule can be deployed in underwater vehicles which requires real-time performance.

\textbf{UnitBackbone.}
Since we aim for less computation and more effective feature information for THead, we build the stem using two successive \(3 \times 3\) convolutions (convs) with \(2 \times\) downsampling. The input image (\(H \times W\)) is transformed by the stem into features with a resolution of \(H/4 \times W/4\). And two depth-wise (DW) large kernel convolution blocks are arranged after the stem. We design the LK Block with reference to RepLKNet~\cite{53}, we use a \(1 \times 1\) conv before and after the DW large kernel convolution layer which uses a \(3 \times 3\) kernel for re-parameterization. This structure captures sufficient receptive field and nonlinear aggregation information~\cite{53}. The UnitBackbone is lightweight, containing only two convs in the stem and two LK blocks. Each of them has architectural hyper-parameters that define the channel dimension C. And there is also the large kernel size K in LK Block. So that the UnitBackbone architecture is defined by [\(C_{s1}\),\(C_{s2}\),\(C_{1}\),\(C_{2}\)], [\(K_{1}\),\(K_{2}\)]. Since the LK Block has the same channel dimension as the output of the stem, \(C_{s2}=C_{1}=C_{2}\). The architecture is simplified as [\(C_{s1}\),\(C_{s2}\)], [\(K_{1}\),\(K_{2}\)].

\textbf{THead.}
The features extracted by the UnitBackbone are used by the THead to estimate the transmission map \(\bm{t}\) in Eq.~\eqref{eq:km}. The THead has two \(3 \times 3\) convs, each of which is preceded by a \(2 \times\) upsampling. The first conv has the same channel as the output of the UnitBackbone. Finally, the estimated transmission map \(\bm{t}\) with 3 channel dimensions is bound to fall between 0 and 1 by the sigmoid activation function. The size and channel dimension of the transmission map is the same as the input image.

\textbf{AHead.}
The global background light \(\bm{A}\) in Eq.~\eqref{eq:km} is calculated by AHead, which is the mean value of each channel in the input image. We use the mean value as \(\bm{A}\) instead of the neural network to estimate it~\cite{24,55}. In this way, the gradient coupling in the UnitBackbone between AHead and THead is removed during training, which makes UnitModule more focused on estimating transmission maps.

UnitModule calculates the enhanced image by using Eq.~\eqref{eq:transform km}. And it is optimized for its loss (in Section~\ref{sec:Loss Functions}) and detection loss. The loss of UnitModule and the formula Eq.~\eqref{eq:transform km} are weak constraints, which just point the way for the UnitModule to learn how to enhance the image. It is the key to ensuring that the UnitModule is fully optimized for detection loss. In other words, make the image enhanced by UnitModule be what the object detector prefers. Instead of letting the UnitModule enhance the image exactly according to the principle of the modified Koschmieder's model.

\subsection{Unsupervised Learning}
\label{sec:Unsupervised Learning}

With reference to~\cite{24}, we design an unsupervised learning strategy for training UnitModule. So that UnitModule can be trained with the detector without any additional dataset. In this work, we only need to consider how the UnitModule learns about the transmission map. The learning strategy is summarized as follows.

\noindent Rewrite the modified Koschmieder's model Eq.~\eqref{eq:km} and further degrade the input image \(\bm{J_{1}}\) using \(\bm{t_{1}}\) and \(\bm{A}\).
\begin{align}
    \label{eq:re km}
    \bm{J_{1}} & = \bm{J}\bm{t_{1}} + (1 - \bm{t_{1}})\bm{A} \\
    \label{eq:degrad input image}
    \bm{J_{2}} & = \bm{J_{1}}\bm{t}' + (1 - \bm{t}')\bm{A}
\end{align}
\noindent From Eq.~\eqref{eq:re km} and Eq.~\eqref{eq:degrad input image}, we get:
\begin{align}
    \label{eq:Middle J12}
    \bm{J_{2}} & = \bm{J}\bm{t_{1}}\bm{t}' + \bm{A}\bm{t}' - \bm{A}\bm{t_{1}}\bm{t}' + \bm{A} - \bm{A}\bm{t}' \\
    \label{eq:J12}
    \bm{J_{2}} & = \bm{J}\bm{t_{1}}\bm{t}' + (1 - \bm{t_{1}}\bm{t}')\bm{A}
\end{align}
where \((x)\) is dropped, and \(\bm{J_{1}}\) is the degraded image (the input image). \(\bm{J}\) is the undegraded image corresponding to \(\bm{J_{1}}\), which is the enhanced image. \(\bm{J_{2}}\) is a further degeneration of \(\bm{J_{1}}\). \(\bm{A}\) is the global background light of \(\bm{J_{1}}\). \(\bm{t_{1}}\) is the transmission map estimated by UnitModule. Setting a hyper-parameter \(\bm{t}'\) to the degraded \(\bm{J_{1}}\) gets \(\bm{J_{2}}\) in Eq.~\eqref{eq:degrad input image}. In theory, enhancing either \(\bm{J_{1}}\) or \(\bm{J_{2}}\) should result in \(\bm{J}\), that is, inputting \(\bm{J_{1}}\) into UnitModule should get \(\bm{t_{1}}\) and inputting \(\bm{J_{2}}\) into UnitModule should get \(\bm{t_{1}}\bm{t}'\). Let \(\alpha=\bm{t}'\), and \(\alpha\) be a hyper-parameter to control the degree of degradation of \(\bm{J_{2}}\), its range is between 0 and 1. Rewrite Eq.~\eqref{eq:J12}, then we get:
\begin{align}
    \label{eq:J12a1}
    \bm{J_{2}} & = \bm{J}\alpha\bm{t_{1}} + (1 - \alpha\bm{t_{1}})\bm{A} \\
    \label{eq:J12a2}
               & = \bm{J_{1}}\alpha + (1 - \alpha)\bm{A}
\end{align}

\begin{figure}[t]
	\centering
		\includegraphics[width=\linewidth]{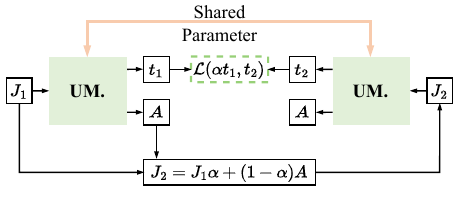}
	\caption{
    The calculation process of transmission map loss for unsupervised learning. \(\bm{J_{1}}\) is the input degraded image and \(\bm{J_{2}}\) is the image degraded from \(\bm{J_{1}}\) using the hyper-parameter \(\alpha\) and the calculated global background light \(\bm{A}\) of \(\bm{J_{1}}\). UM. indicates the UnitModule.
    }
	\label{fig:loss t}
\end{figure}

According to the unsupervised learning strategy, we get the transmission map loss. Its detailed calculation process is shown in Figure~\ref{fig:loss t}. We calculate \(\bm{J_{2}}\) from \(\bm{J_{1}}\), and obtain the transmission map \(\bm{t_{2}}\) and global background light \(\bm{A}\) for \(\bm{J_{2}}\) using the shared-parameter UnitModule. Then we optimize the UnitModule by minimizing the dissimilarity between \(\alpha\bm{t_{1}}\) and \(\bm{t_{2}}\), which the detailed loss function is described in Section~\ref{sec:Loss Functions}.

\begin{figure*}[t]
	\centering
		\includegraphics[width=\linewidth]{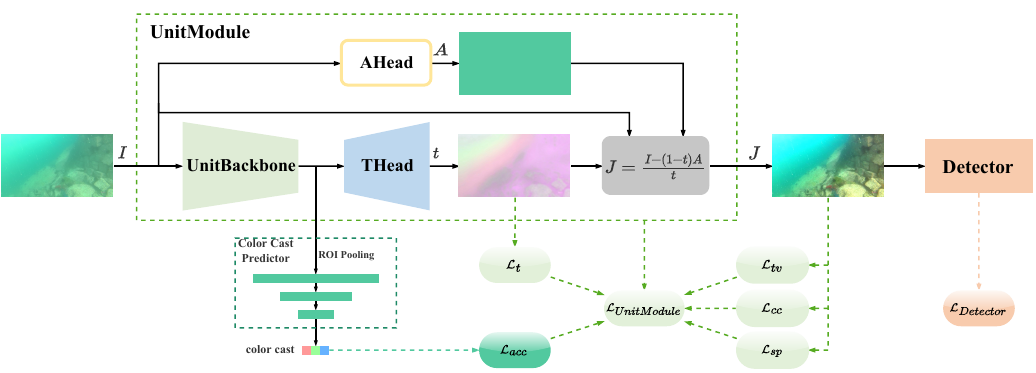}
	\caption{
    Joint training framework of the detector with UnitModule. UnitModule is based on the modified Koschmieder’s model~\cite{23} to enhance the input image (the enhanced input image is a tensor which requires that its gradient can be backpropagated between UnitModule and Detector). The dotted box shows the detailed inference flow of UnitModule, and the dotted line points to the object of the loss optimization.
    }
	\label{fig:joint training}
\end{figure*}

It is noticed that the \(\bm{A}\) of \(\bm{J_{1}}\) and \(\bm{J_{2}}\) are exactly the same, which is the key to implementing transmission map loss. The relevant derivation is as follows:
\begin{align}
    \label{eq:J1A}
    \mu_{c}(\bm{J_{1}}) & = \bm{A_{1}} = \bm{A} \\
    \label{eq:J2A}
    \mu_{c}(\bm{J_{2}}) & = \bm{A_{2}} \\
                        & = \mu_{c}\left(\bm{J_{1}}\alpha + (1-\alpha)\bm{A_{1}}\right) \\
                        & = \mu_{c}(\bm{J_{1}})\alpha + (1-\alpha)\bm{A_{1}} \\
                        & = \bm{A_{1}} = \bm{A}
\end{align}
where \(\mu_{c}(*)\) is the mean of the image in the channel dimension, \(\bm{A_{1}}\) and \(\bm{A_{2}}\) are the calculated global background light of \(\bm{J_{1}}\) and \(\bm{J_{2}}\) respectively.

\subsection{Loss Functions}
\label{sec:Loss Functions}

We design several loss functions for UnitModule based on some assumptions and priors, and we design a color cast predictor to assist UnitModule in gaining a better understanding of color noise in underwater images. At the same time, UnitModule is also optimized for object detection loss functions.

\textbf{Transmission Map Loss.}
The transmission map loss \(\mathcal{L}_{t}\) plays an important role in the training of UnitModule. It improves the performance of UnitModule in estimating the pixel-wise information in the transmission map. According to Section~\ref{sec:Unsupervised Learning}, the loss is formulated as:
\begin{equation}
    \label{eq:loss t}
    \mathcal{L}_{t} = \sum_{x}
    \left\| \alpha\hat{\bm{t_{1}}}(x) - \hat{\bm{t_{2}}}(x) \right\|_{2}^{2}
\end{equation}
where \(x\) is a pixel. \(\hat{\bm{t_{1}}}(x)\) and \(\hat{\bm{t_{2}}}(x)\) are estimated by UnitModule based on \(\bm{J_{1}}\) and \(\bm{J_{2}}\), respectively.

\textbf{Saturated Pixel Loss.}
The pixel value of image \(\hat{\bm{J}}\) enhanced by the UnitModule might exceed the range of normalized images that is [0,1]. Using saturated pixel loss \(\mathcal{L}_{sp}\) restricts the pixel value of \(\hat{\bm{J}}\) to alleviate overflow/underflow that results in pixel value saturation, so that the gradient flow will not be stopped by the clipping operation. It is defined as:
\begin{equation}
    \label{eq:loss sp}
    \begin{split}
    \mathcal{L}_{sp}
    & = \sum_{x} \left(\max(\hat{\bm{J}}(x), 1) + \max(\hat{\bm{J}}'(x), 1) \right) \\
    & - \sum_{x} \left(\min(\hat{\bm{J}}(x), 0) + \min(\hat{\bm{J}}'(x), 0) \right)
    \end{split}
\end{equation}
where \(x\) is a pixel, \(\hat{\bm{J}}\) and \(\hat{\bm{J}}'\) are the enhanced image from \(\bm{J_{1}}\) and \(\bm{J_{2}}\) respectively.

\textbf{Total Variation Loss.}
The total variation loss \(\mathcal{L}_{tv}\) is used to improve the performance of noise reduction for UnitModule and the robustness to noise in image enhancement. It makes the enhanced image more smooth. This loss is calculated as:
\begin{equation}
    \label{eq:loss tv}
    \begin{split}
    \mathcal{L}_{tv} 
    & = \sum_{h} \left\| \hat{\bm{J}}(h + 1) - \hat{\bm{J}}(h) \right\|_{2}^{2} \\
    & + \sum_{w} \left\| \hat{\bm{J}}(w + 1) - \hat{\bm{J}}(w) \right\|_{2}^{2}
    \end{split}
\end{equation}
where \(h\) and \(w\) represent the vertical and horizontal pixel coordinates of the image respectively.

\textbf{Color Cast Loss.}
Since the color cast is common in underwater images, we use the color cast loss \(\mathcal{L}_{cc}\) to guide the UnitModule to learn how to relieve color cast. The color cast loss is based on the gray-word assumption. The formula is as:
\begin{equation}
    \label{eq:loss cc}
    \mathcal{L}_{cc} = \sum_{(c_{1}, c_{2}) \in \Delta}
    \left\| \mu(\hat{\bm{J}}^{c_{1}}) - \mu(\hat{\bm{J}}^{c_{2}}) \right\|_{2}^{2}
\end{equation}
where \(c\) is the image color channel, \(\mu(\hat{\bm{J}}^{c_{x}})\) is the mean of the enhanced image in \(c_{x}\), \(\Delta = \left\{ (\text{R},\text{G}),(\text{G},\text{B}),(\text{B},\text{R}) \right\}\) is a set containing RGB color channel pairs.

\begin{figure*}[t]
	\centering
		\includegraphics[width=\linewidth]{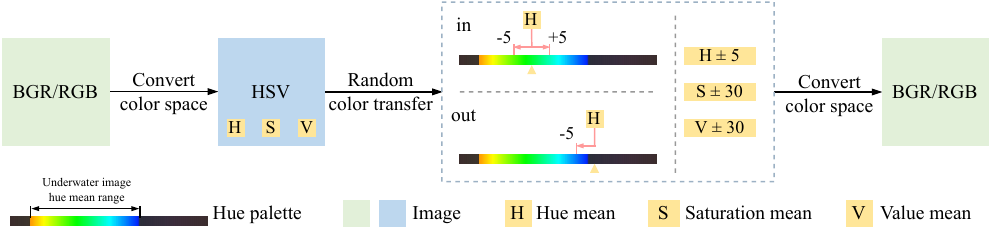}
	\caption{
    The implementation process of UCRT. The H, S, and V are added or subtracted by a random number in the specified range. The in/out (in dotted box) indicates that the hue mean is in/out of the underwater image hue mean range [18, 116]. The range of the hue palette is [0, 180].
    }
	\label{fig:UCRT}
\end{figure*}
\begin{table}[!t]
\centering

\begin{tabular}{@{}lcc@{}}
\toprule
Dataset & Hue min & Hue max \\ 
\midrule

DUO~\cite{25}  & 18.76 & 95.48  \\
URPC2020       & 17.97 & 99.64  \\
URPC2021       & 17.97 & 103.24 \\
UIEB~\cite{68} & 25.54 & 116.34 \\ \midrule
Hue range      & 18    & 116    \\

\bottomrule
\end{tabular}
    \caption{
    The hue mean range in different underwater datasets.
    }
    \label{table:hue range}
\end{table}

\textbf{Assisting Color Cast Loss.}
We design a color cast predictor to predict the color cast of underwater images. It can assist UnitBackbone in better understanding the color cast of underwater images and improve the prediction ability of UnitModule for the transmission map while being removed during the model inference process. It consists mainly of Region of interest pooling~\cite{28} (ROI pooling) and linear layers. Features from UnitBackbone are transformed into 7 \(\times\) 7 size features by the ROI pooling layer and then the 1 \(\times\) 1 conv transforms the feature channels to 3. Then the color cast of underwater images is calculated by three linear layers with 32, 16, and 1 channels. Except for the last layer, all other layers have the following GN and ReLU. The last layer has the following sigmoid activation function to limit the range of the color cast value between 0 and 1. The predicted color cast has values corresponding to the RGB three channels of the image, and then we use the global background light \(\bm{A}\) as the color cast label to calculate the assisting color cast loss \(\mathcal{L}_{acc}\). The loss is formulated as:
\begin{equation}
    \label{eq:loss acc}
    \mathcal{L}_{acc} = \left\| \hat{\bm{C}} - \bm{A} \right\|_{2}^{2}
\end{equation}
where \(\hat{\bm{C}}\) is the predicted color cast.

In our joint training in Figure~\ref{fig:joint training}, UnitModule is optimized for the UnitModule loss \(\mathcal{L}_{UnitModule}\) and the object detection loss \(\mathcal{L}_{Detector}\) which varies with different detectors. The UnitModule loss \(\mathcal{L}_{UnitModule}\) consists of \(\mathcal{L}_{t}\), \(\mathcal{L}_{sp}\), \(\mathcal{L}_{tv}\), \(\mathcal{L}_{cc}\) and \(\mathcal{L}_{acc}\). The total loss function \(\mathcal{L}\) for the detector with the UnitModule is:
\begin{align}
    \label{eq:UnitModule loss}
    \mathcal{L}_{UnitModule} = & w_{1}\mathcal{L}_{t}+w_{2}\mathcal{L}_{sp}+w_{3}\mathcal{L}_{tv} \\
                               & +w_{4}\mathcal{L}_{cc}+w_{5}\mathcal{L}_{acc} \\
    \label{eq:all loss}
    \mathcal{L} = & \mathcal{L}_{UnitModule} + \mathcal{L}_{Detector}
\end{align}
The configuration of weight \(w_{x}\) is described in~\ref{appendix:loss weights}.

\subsection{Underwater Image Data Augmentation}
\label{sec:Underwater Image Data Augmentation}

We design a data augmentation called Underwater Color Random Transfer (UCRT) to convert the color cast of underwater images with reference to the HSV augmentation code in~\cite{38}. In this method, the input image is converted to the HSV color space, while the color cast of images is mainly dominated by the mean of H (Hue). We calculated the minimum and maximum of the mean of H on four underwater image datasets (DUO~\cite{25}, URPC2020, URPC2021, and UIEB~\cite{68}), as shown in Table~\ref{table:hue range}. To keep the image within the underwater color cast range, its mean of H should be limited to this range [18,116], which is obtained on a smaller range by rounding the mean of H in the four ranges above.

To transfer the color cast of the image, we mainly limit the hue mean of the image in HSV color space, as shown in Figure~\ref{fig:UCRT}. First, we calculate the hue mean for the HSV color space image and determine whether it is within the hue mean range of the underwater image. If the hue mean is within this range, a random number in the range [-5, +5] is added to the H channel of the image while keeping the hue mean within the hue mean range of the underwater image. Otherwise, the H channel of the image is randomly shifted within a distance of 5 in the direction of the hue mean range of the underwater image. A random number in the range [-30, +30] is added to the S and V channels of the image to randomly change the saturation and value. The implementation probabilities for H, S, and V are all 0.5.
\section{Experiments}
\label{sec:Experiments}

\begin{table*}[t]
\centering

\resizebox{\linewidth}{!}{
\begin{tabular}{@{}lllccccccccc@{}}
\toprule
  Method &
  \(K_{1}\)-\(K_{2}\) &
  \(C_{s1}\)-\(C_{s2}\) &
  Params(M)\(\downarrow\) &
  FLOPs(G)\(\downarrow\) &
  FPS\(\uparrow\) &
  AP\(\uparrow\) &
  \(\text{AP}_{50}\)\(\uparrow\) &
  \(\text{AP}_{75}\)\(\uparrow\) &
  \(\text{AP}_{S}\)\(\uparrow\) &
  \(\text{AP}_{M}\)\(\uparrow\) &
  \(\text{AP}_{L}\)\(\uparrow\) \\
\midrule

YOLOX-S~\cite{38} & -      & -      &  8.939 & 13.32  & 79.3 & 61.3                & 82.0          & 68.6          & 52.5          & 62.9          & 59.9          \\
w/ UnitModule     & 5--5   & 32--64 & +0.080 & +5.69  & 76.1 & 61.6                & 83.7          & 68.9          & 52.7          & 62.4          & 61.0          \\
w/ UnitModule     & 7--7   & 32--64 & +0.083 & +5.77  & 75.4 & 62.6                & 85.2          & 69.8          & \textbf{58.1} & \textbf{64.9} & 60.7          \\
w/ UnitModule     & 9--9   & 32--64 & +0.087 & +5.88  & 74.8 & \textbf{63.2(+1.9)} & \textbf{85.7} & \textbf{70.0} & 55.6          & 64.4          & \textbf{62.2} \\
w/ UnitModule     & 11--11 & 32--64 & +0.093 & +6.01  & 74.3 & 62.4                & 84.7          & \textbf{70.0} & 56.1          & 63.5          & 61.4          \\ \midrule
w/ UnitModule     & 9--9   & 32--16 & +0.013 & +0.76  & 76.2 & 62.9                & 84.7          & 70.3          & \textbf{59.3} & 64.1          & 61.9          \\
w/ UnitModule     & 9--9   & 32--32 & +0.031 & +1.94  & 75.9 & \textbf{63.7(+2.4)} & \textbf{85.8} & \textbf{72.2} & 58.6          & \textbf{65.1} & \textbf{62.3} \\

\bottomrule
\end{tabular}}
    \caption{
    Comparison between the different architectures of UnitModule on DUO~\cite{25}. FLOPs are measured using the input image size of 640 \(\times\) 640 and FPS is measured using a  NVIDIA 3090Ti GPU. + indicates that adding its item to the YOLOX-S line.
    }
    \label{table:architecture}
\end{table*}
\begin{table}[t]
\centering

\begin{tabular}{@{}ccccc@{}}
\toprule
UnitModule & UCRT & AP\(\uparrow\) & \(\text{AP}_{50}\)\(\uparrow\) & \(\text{AP}_{75}\)\(\uparrow\) \\ 
\midrule

           &            & 61.3          & 82.0          & 68.6          \\
\checkmark &            & 62.4          & 84.1          & 70.9          \\
           & \checkmark & 62.0          & 83.9          & 70.5          \\
\checkmark & \checkmark & \textbf{63.7} & \textbf{85.8} & \textbf{72.2} \\

\bottomrule
\end{tabular}
    \caption{
    Ablation study of UnitModule and UCRT.
    }
    \label{table:unitmodule and ucrt}
\end{table}
\begin{table}[t]
\centering

\begin{tabular}{@{}lccc@{}}
\toprule
Loss & AP\(\uparrow\) & \(\text{AP}_{50}\)\(\uparrow\) & \(\text{AP}_{75}\)\(\uparrow\) \\ 
\midrule

only \(\mathcal{L}_{t}\)          & 62.1          & 83.9          & 70.3          \\
w/o  \(\mathcal{L}_{sp}\)         & 63.5          & 85.5          & 71.9          \\
w/o  \(\mathcal{L}_{tv}\)         & 63.1          & 85.0          & 71.4          \\
w/o  \(\mathcal{L}_{cc}\)         & 63.0          & 85.1          & 71.2          \\
w/o  \(\mathcal{L}_{acc}\)        & 62.7          & 84.5          & 70.7          \\
all  \(\mathcal{L}_{UnitModule}\) & \textbf{63.7} & \textbf{85.8} & \textbf{72.2} \\

\bottomrule
\end{tabular}
    \caption{
    Ablation study of losses for UnitModule.
    }
    \label{table:losses}
\end{table}

\textbf{Dataset and evaluation metrics.}
We conduct experiments on the underwater dataset DUO~\cite{25}, which includes the holothurian, echinus, scallop, and starfish classes. The DUO dataset consists of the training set (6671 images) and the test set (1111 images) for training and validation, respectively. We combine the URPC2020 and URPC2021 datasets and use the Perceptual Hash algorithm (PHash) to remove similar or repeated images in the DUO dataset. After deduplication, we get 1042 images and call them \(\text{URPC}_{test}\) as the test set to verify the performance of the detection model on the new test dataset. The performance is measured by COCO~\cite{79} Average Precision (AP).

\textbf{Implementation details.}
Our implementations are based on the MMDetection toolbox~\cite{60}. All object detection models are trained on 2 NVIDIA A6000 GPUs. For ablation studies, we train YOLOX-S~\cite{38} with UnitModule for 100 epochs and use only the horizontal RandomFlip data augmentation for the 640 \(\times\) 640 input image. When jointly training an object detection model with UnitModule, we do not need to change any of the original configuration parameters of the object detection model, we just need to insert the UnitModule. For UnitModule, we set a hyper-parameter \(t_{min}=0.001\) to limit the minimum value per pixel in the predicted transmission map and reduce the generation of distorted images. When joint training with different object detection models, the weight of UnitModule loss is different, we set different UnitModule loss weights to keep the UnitModule loss value and detector loss value in the same order of magnitude. The detailed training configuration of different object detection models is described in~\ref{appendix:training configuration}.

\subsection{Ablation Study}
\label{sec:Ablation Study}

\begin{table}[t]
\centering

\begin{tabular}{@{}cccc@{}}
\toprule
\(\alpha\) & AP\(\uparrow\) & \(\text{AP}_{50}\)\(\uparrow\) & \(\text{AP}_{75}\)\(\uparrow\) \\ 
\midrule

0.85          & 63.4          & 85.7          & 72.0          \\
\textbf{0.90} & \textbf{63.7} & \textbf{85.8} & 72.2          \\
0.95          & 63.1          & 85.1          & \textbf{72.8} \\

\bottomrule
\end{tabular}
    \caption{
    Ablation study of the hyper-parameter \(\alpha\).
    }
    \label{table:alpha}
\end{table}
\begin{table}[t]
\centering

\begin{tabular}{@{}lccc@{}}
\toprule
Method & AP\(\uparrow\) & \(\text{AP}_{50}\)\(\uparrow\) & \(\text{AP}_{75}\)\(\uparrow\) \\ 
\midrule

YOLOX-S~\cite{38}                  & 61.3                & 82.0          & 68.6          \\ \midrule
w/ UDCP~\cite{71}                  & 59.5(-1.8)          & 81.5          & 65.4          \\
w/ RGHS~\cite{72}                  & 59.6(-1.7)          & 83.2          & 64.8          \\
w/ Song \textit{et al.}~\cite{50}  & 61.2(-0.1)          & 83.4          & 67.9          \\
w/ WaterNet~\cite{68}              & 60.0(-1.3)          & 82.0          & 66.6          \\
w/ \(\text{FUnIE\_GAN}\)~\cite{75} & 58.3(-3.0)          & 79.9          & 64.1          \\
w/ \(\text{U\_shape}\)~\cite{76}   & 53.3(-8.0)          & 73.9          & 57.6          \\
w/ UnitModule (Ours)               & \textbf{63.7(+2.4)} & \textbf{85.8} & \textbf{72.2} \\

\bottomrule
\end{tabular}
    \caption{
    Comparison of YOLOX-S trained with different underwater image enhancement methods on DUO~\cite{25}.
    }
    \label{table:comparison of image enhancement}
\end{table}

\textbf{UnitModule architecture.}
We compare YOLOX-S~\cite{38}, which inserted the UnitModule with different architecture in Table~\ref{table:architecture}. We first compare the effectiveness of different large kernel sizes in the LK Block, with the kernel sizes ranging from 5 \(\times\) 5 to 11 \(\times\) 11., which peaked at 63.2 AP at kernel size 9 \(\times\) 9. Then we compare the channel dimension of the stem in UnitModule from 16 to 64 and get that the UnitModule with K=[9,9], C=[32,32] has the best performance, outperforming the YOLOX-S without UnitModule (by 2.4 AP). As a result, we choose this configuration as the best for UnitModule. And this plug-and-play UnitModule has a small number of parameters of 0.031M and a low computing complexity of 1.94G FLOPs, while it only reduces the FPS from 79.3 to 75.9 in YOLOX-S, which has little impact on the inference speed of the original detection model.

\begin{table*}[!t]
\centering

\begin{tabular}{@{}lcccccccc@{}}
\toprule
  \multirow{2}{*}{Method} &
  \multirow{2}{*}{Params(M)\(\downarrow\)} &
  \multirow{2}{*}{FLOPs(G)\(\downarrow\)} &
  \multicolumn{3}{c}{DUO} &
  \multicolumn{3}{c}{\(\text{URPC}_{test}\)} \\ \cmidrule(l){4-6}\cmidrule(l){7-9}  
  & & &
  AP\(\uparrow\) &
  \(\text{AP}_{50}\)\(\uparrow\) &
  \(\text{AP}_{75}\)\(\uparrow\) &
  AP\(\uparrow\) &
  \(\text{AP}_{50}\)\(\uparrow\) &
  \(\text{AP}_{75}\)\(\uparrow\) \\
\midrule

YOLOv5-S~\cite{36} & 7.03  & 7.94  & 46.0                & 70.1          & 51.8          & 26.0                & 54.2          & 21.2          \\
w/ UnitModule      & 7.06  & 9.88  & \textbf{48.6(+2.6)} & \textbf{73.2} & \textbf{55.7} & \textbf{29.3(+3.3)} & \textbf{59.4} & \textbf{25.3} \\
YOLOv5-M           & 20.88 & 24.04 & 59.0                & 78.7          & 67.4          & 33.1                & 62.5          & 31.3          \\
w/ UnitModule      & 20.91 & 25.98 & \textbf{60.8(+1.8)} & \textbf{81.3} & \textbf{69.9} & \textbf{35.6(+2.5)} & \textbf{66.5} & \textbf{34.6} \\
YOLOv5-L           & 46.15 & 53.99 & 61.0                & 80.0          & 68.9          & 35.0                & 64.7          & 34.2          \\
w/ UnitModule      & 46.19 & 55.93 & \textbf{61.8(+0.8)} & \textbf{82.0} & \textbf{70.4} & \textbf{36.3(+1.3)} & \textbf{67.4} & \textbf{36.2} \\ \midrule
YOLOv6-S~\cite{37} & 18.84 & 24.20 & 58.5                & 78.5          & 66.1          & 33.5                & 62.9          & 32.0          \\
w/ UnitModule      & 18.87 & 26.14 & \textbf{60.7(+2.2)} & \textbf{82.5} & \textbf{69.6} & \textbf{36.3(+2.8)} & \textbf{68.8} & \textbf{35.1} \\
YOLOv6-M           & 37.08 & 44.42 & 60.8                & 80.4          & 68.2          & 34.5                & 64.6          & 32.8          \\ 
w/ UnitModule      & 37.11 & 46.36 & \textbf{62.2(+1.4)} & \textbf{83.5} & \textbf{70.1} & \textbf{36.6(+2.1)} & \textbf{69.1} & \textbf{35.7} \\
YOLOv6-L           & 58.46 & 71.32 & 62.9                & 82.0          & 70.5          & 35.5                & 66.0          & 34.7          \\
w/ UnitModule      & 58.49 & 73.26 & \textbf{63.7(+0.8)} & \textbf{83.3} & \textbf{72.1} & \textbf{36.8(+1.3)} & \textbf{68.0} & \textbf{36.6} \\ \midrule
YOLOv7-T~\cite{81} & 6.02  & 6.56  & 26.9                & 45.7          & 28.5          & 14.5                & 33.7          & 9.5           \\
w/ UnitModule      & 6.05  & 8.50  & \textbf{28.6(+1.7)} & \textbf{48.5} & \textbf{30.6} & \textbf{16.9(+2.4)} & \textbf{37.3} & \textbf{13.5} \\
YOLOv7-L           & 37.21 & 52.41 & 35.7                & 56.6          & 39.1          & 20.3                & 44.6          & 15.3          \\
w/ UnitModule      & 37.24 & 54.35 & \textbf{36.4(+0.7)} & \textbf{58.1} & \textbf{40.5} & \textbf{21.8(+1.5)} & \textbf{47.8} & \textbf{17.6} \\ \midrule
YOLOv8-S~\cite{82} & 11.14 & 14.27 & 60.9                & 80.2          & 68.1          & 34.6                & 63.8          & 33.9          \\
w/ UnitModule      & 11.17 & 16.21 & \textbf{63.2(+2.3)} & \textbf{83.2} & \textbf{71.8} & \textbf{37.5(+2.9)} & \textbf{69.4} & \textbf{37.0} \\
YOLOv8-M           & 25.86 & 39.44 & 63.0                & 81.7          & 69.9          & 36.5                & 65.6          & 36.7          \\
w/ UnitModule      & 25.89 & 41.38 & \textbf{64.6(+1.6)} & \textbf{84.8} & \textbf{72.2} & \textbf{38.6(+2.1)} & \textbf{69.5} & \textbf{39.2} \\
YOLOv8-L           & 43.63 & 82.56 & 64.0                & 82.2          & 72.0          & 37.0                & 66.1          & 37.6          \\
w/ UnitModule      & 43.66 & 84.50 & \textbf{65.1(+1.1)} & \textbf{84.1} & \textbf{74.1} & \textbf{38.3(+1.3)} & \textbf{68.1} & \textbf{39.7} \\ \midrule
YOLOX-S~\cite{38}  & 8.94  & 13.32 & 61.3                & 82.0          & 68.6          & 36.5                & 68.2          & 35.0          \\
w/ UnitModule      & 8.97  & 15.26 & \textbf{63.7(+2.4)} & \textbf{85.8} & \textbf{72.2} & \textbf{39.6(+3.1)} & \textbf{73.8} & \textbf{38.4} \\
YOLOX-M            & 25.28 & 36.76 & 64.6                & 83.3          & 72.1          & 38.6                & 69.5          & 38.7          \\
w/ UnitModule      & 25.31 & 38.70 & \textbf{66.4(+1.8)} & \textbf{86.5} & \textbf{74.6} & \textbf{41.1(+2.5)} & \textbf{72.9} & \textbf{42.3} \\ 
YOLOX-L            & 54.15 & 77.66 & 66.2                & 84.8          & 73.9          & 38.8                & 70.2          & 39.1          \\
w/ UnitModule      & 54.18 & 79.60 & \textbf{67.1(+0.9)} & \textbf{86.0} & \textbf{76.0} & \textbf{40.3(+1.5)} & \textbf{73.3} & \textbf{41.6} \\ \midrule
RTMDet-S~\cite{83} & 8.86  & 14.75 & 63.4                & 82.6          & 70.8          & 37.2                & 68.8          & 36.1          \\
w/ UnitModule      & 8.89  & 16.69 & \textbf{65.3(+1.9)} & \textbf{86.7} & \textbf{73.7} & \textbf{39.9(+2.7)} & \textbf{73.3} & \textbf{39.2} \\
RTMDet-M           & 24.67 & 39.08 & 63.8                & 83.2          & 72.2          & 37.3                & 68.5          & 37.1          \\
w/ UnitModule      & 24.70 & 41.02 & \textbf{64.9(+1.1)} & \textbf{85.2} & \textbf{74.0} & \textbf{39.3(+2.0)} & \textbf{72.3} & \textbf{39.8} \\
RTMDet-L           & 52.26 & 79.96 & 63.8                & 83.2          & 71.8          & 37.5                & 68.9          & 36.6          \\
w/ UnitModule      & 52.29 & 81.90 & \textbf{64.4(+0.6)} & \textbf{85.0} & \textbf{73.2} & \textbf{39.0(+1.5)} & \textbf{71.2} & \textbf{39.2} \\

\bottomrule
\end{tabular}
    \caption{
    Comparison of different YOLO-like object detection models with UnitModule on DUO~\cite{25} and \(\text{URPC}_{test}\) test set. FLOPs are measured on the input image size of 640 \(\times\) 640.
    }
    \label{table:obyolo}
\end{table*}

\textbf{UnitModule and UCRT data augmentation.}
We conduct ablation studies based on YOLOX-S~\cite{38} for UnitModule and UCRT data augmentation to demonstrate the effectiveness of the UnitModule and UCRT. The results in Table~\ref{table:unitmodule and ucrt} indicate that UnitModule and UCRT each improve the performance of the object detection model on the underwater dataset by 1.1 AP and 0.7 AP, respectively. While UCRT leads to a further 1.3 increase in AP compared with the model with only the UnitModule. The results show that UCRT effectively improves the performance of UnitModule. Unless specific, We use UCRT in the training of UnitModule.

\textbf{UnitModule losses and a hyper-parameter \(\alpha\).}
In Table~\ref{table:losses}, we compare different loss functions on the performance of UnitModule. The results indicate that each loss function contributes to the training of UnitModule and that the assisting color cast loss has the greatest impact, while the saturated pixel loss has the smallest impact. For the hyper-parameter \(\alpha\), in Table~\ref{table:alpha}, the setting of \(\alpha\) to 0.90 achieves the optimal performance of object detection.

\begin{table*}[!th]
\centering
\resizebox{\linewidth}{!}{

\begin{tabular}{@{}lccccccccc@{}}
\toprule
  \multirow{2}{*}{Method} &
  \multirow{2}{*}{Backbone} &
  \multirow{2}{*}{Params(M)\(\downarrow\)} &
  \multirow{2}{*}{FLOPs(G)\(\downarrow\)} &
  \multicolumn{3}{c}{DUO} &
  \multicolumn{3}{c}{\(\text{URPC}_{test}\)} \\ \cmidrule(l){5-7}\cmidrule(l){8-10}  
  & & & &
  AP\(\uparrow\) &
  \(\text{AP}_{50}\)\(\uparrow\) &
  \(\text{AP}_{75}\)\(\uparrow\) &
  AP\(\uparrow\) &
  \(\text{AP}_{50}\)\(\uparrow\) &
  \(\text{AP}_{75}\)\(\uparrow\) \\
\midrule

Faster R-CNN~\cite{28}  & ResNet-50 & 41.36 & 208.00 & 63.5                & 83.6          & 71.5          & 40.4                & 71.9          & 41.3          \\
w/ UnitModule           & ResNet-50 & 41.40 & 209.94 & \textbf{64.9(+1.4)} & \textbf{86.6} & \textbf{73.2} & \textbf{42.7(+2.3)} & \textbf{77.0} & \textbf{43.4} \\ \midrule
Cascade R-CNN~\cite{63} & ResNet-50 & 69.16 & 236.00 & 64.8                & 83.5          & 73.0          & 40.7                & 72.5          & 42.1          \\
w/ UnitModule           & ResNet-50 & 69.19 & 237.94 & \textbf{65.8(+1.0)} & \textbf{85.6} & \textbf{73.8} & \textbf{42.6(+1.9)} & \textbf{76.2} & \textbf{44.1} \\ \midrule
FCOS~\cite{41}          & ResNet-50 & 32.12 & 198.00 & 62.5                & 82.8          & 70.0          & 39.2                & 72.1          & 38.8          \\
w/ UnitModule           & ResNet-50 & 32.15 & 199.94 & \textbf{64.0(+1.5)} & \textbf{85.6} & \textbf{71.9} & \textbf{41.8(+2.6)} & \textbf{77.6} & \textbf{41.6} \\ \midrule
RetinaNet~\cite{40}     & ResNet-50 & 36.39 & 207.00 & 61.9                & 81.7          & 69.3          & 39.3                & 71.8          & 39.8          \\
w/ UnitModule           & ResNet-50 & 36.42 & 208.94 & \textbf{62.7(+0.8)} & \textbf{82.3} & \textbf{70.6} & \textbf{41.8(+2.5)} & \textbf{76.7} & \textbf{42.8} \\ \midrule
TOOD~\cite{42}          & ResNet-50 & 32.03 & 199.00 & 67.7                & 86.1          & 74.8          & 41.4                & 74.0          & 41.8          \\
w/ UnitModule           & ResNet-50 & 32.06 & 200.94 & \textbf{68.7(+1.0)} & \textbf{87.9} & \textbf{76.3} & \textbf{43.4(+2.0)} & \textbf{76.8} & \textbf{44.7} \\ \midrule
DETR~\cite{43}          & ResNet-50 & 41.56 & 96.51  & 54.5                & 76.4          & 62.8          & 35.0                & 65.1          & 34.7          \\
w/ UnitModule           & ResNet-50 & 41.59 & 98.45  & \textbf{55.4(+0.9)} & \textbf{78.0} & \textbf{64.3} & \textbf{37.3(+2.3)} & \textbf{69.5} & \textbf{37.9} \\ \midrule
DINO~\cite{84}          & ResNet-50 & 47.55 & 274.00 & 65.6                & 84.1          & 72.1          & 41.3                & 73.6          & 41.8          \\
w/ UnitModule           & ResNet-50 & 47.58 & 275.94 & \textbf{66.7(+1.1)} & \textbf{86.4} & \textbf{73.5} & \textbf{43.1(+1.8)} & \textbf{77.0} & \textbf{44.2} \\

\bottomrule
\end{tabular}}
    \caption{
    Comparison of different object detection models with UnitModule on DUO~\cite{25} and \(\text{URPC}_{test}\) test set. The backbone of all models is ResNet-50~\cite{64} pretrained on ImageNet~\cite{65}. FLOPs are measured on the input image size of 1344 \(\times\) 678.
    }
    \label{table:obr50}
\end{table*}
\begin{table*}[!th]
\centering
\resizebox{\linewidth}{!}{

\begin{tabular}{@{}lccccccccc@{}}
\toprule
  \multirow{2}{*}{Method} &
  \multirow{2}{*}{Backbone} &
  \multirow{2}{*}{Params(M)\(\downarrow\)} &
  \multirow{2}{*}{FLOPs(G)\(\downarrow\)} &
  \multicolumn{3}{c}{DUO} &
  \multicolumn{3}{c}{\(\text{URPC}_{test}\)} \\ \cmidrule(l){5-7}\cmidrule(l){8-10}  
  & & & &
  AP\(\uparrow\) &
  \(\text{AP}_{50}\)\(\uparrow\) &
  \(\text{AP}_{75}\)\(\uparrow\) &
  AP\(\uparrow\) &
  \(\text{AP}_{50}\)\(\uparrow\) &
  \(\text{AP}_{75}\)\(\uparrow\) \\
\midrule

Faster R-CNN~\cite{28}  & ResNet-101 & 60.36 & 285.00 & 64.7                & 84.7          & 72.7          & 40.0                & 71.4          & 41.4          \\
w/ UnitModule           & ResNet-101 & 60.39 & 286.94 & \textbf{65.6(+0.9)} & \textbf{86.4} & \textbf{74.0} & \textbf{41.7(+1.7)} & \textbf{75.4} & \textbf{42.8} \\ \midrule
Cascade R-CNN~\cite{63} & ResNet-101 & 88.15 & 312.00 & 65.6                & 83.9          & 73.3          & 40.6                & 71.6          & 42.3          \\
w/ UnitModule           & ResNet-101 & 88.18 & 313.94 & \textbf{66.2(+0.6)} & \textbf{85.3} & \textbf{74.1} & \textbf{41.9(+1.3)} & \textbf{74.0} & \textbf{43.6} \\ \midrule
FCOS~\cite{41}          & ResNet-101 & 51.11 & 275.00 & 64.4                & 84.3          & 71.4          & 39.5                & 72.3          & 39.0          \\
w/ UnitModule           & ResNet-101 & 51.14 & 276.94 & \textbf{65.4(+1.0)} & \textbf{86.6} & \textbf{72.9} & \textbf{41.4(+1.9)} & \textbf{76.6} & \textbf{41.2} \\ \midrule
RetinaNet~\cite{40}     & ResNet-101 & 55.38 & 283.00 & 63.6                & 82.8          & 71.5          & 40.5                & 72.7          & 40.9          \\
w/ UnitModule           & ResNet-101 & 55.42 & 284.94 & \textbf{64.0(+0.4)} & \textbf{83.7} & \textbf{72.0} & \textbf{42.3(+1.8)} & \textbf{76.6} & \textbf{42.9} \\ \midrule
TOOD~\cite{42}          & ResNet-101 & 51.02 & 275.00 & 68.6                & 87.0          & 75.1          & 41.5                & 74.4          & 42.4          \\
w/ UnitModule           & ResNet-101 & 51.05 & 276.94 & \textbf{69.3(+0.7)} & \textbf{88.5} & \textbf{76.1} & \textbf{42.9(+1.4)} & \textbf{76.7} & \textbf{44.5} \\ \midrule
DETR~\cite{43}          & ResNet-101 & 60.55 & 171.00 & 55.7                & 77.6          & 64.4          & 35.8                & 65.0          & 36.5          \\
w/ UnitModule           & ResNet-101 & 60.58 & 172.94 & \textbf{56.2(+0.5)} & \textbf{78.6} & \textbf{65.1} & \textbf{37.3(+1.5)} & \textbf{67.8} & \textbf{38.2} \\

\bottomrule
\end{tabular}}
    \caption{
    Comparison of different object detection models with UnitModule on DUO~\cite{25} and \(\text{URPC}_{test}\) test set. The backbone of all models is ResNet-101~\cite{64} pretrained on ImageNet~\cite{65}. FLOPs are measured on the input image size of 1344 \(\times\) 678.
    }
    \label{table:obr101}
\end{table*}

\subsection{Comparison of UnitModule}
\label{sec:Comparison of UnitModule}

We compare the performance of our UnitModule with other underwater image enhancement methods by training YOLOX-S~\cite{38} on images processed by these methods, as shown in Table~\ref{table:comparison of image enhancement}. Our UnitModule achieves the highest performance improvement of 2.4 AP, outperforming the other methods. The performance of the other methods drops 0.1-8.0 AP. These methods which have no interaction with the detector not only do not improve the performance of the detector, they actually decrease it.

\subsection{Performance of UnitModule}
\label{sec:Performance of UnitModule}

\textbf{Performance on YOLO-like object detectors.}
As demonstrated in Table~\ref{table:obyolo}, we compare the performance of different YOLO-like object detectors with UnitModule on DUO~\cite{25} and \(\text{URPC}_{test}\) test set. All models are only trained on the DUO training set and are tested on the DUO test set and \(\text{URPC}_{test}\) test set. Jointly training with UnitModule has improved the performance of object detection models at different parameter scales (T, S, M, L). UnitModule achieves the greatest performance improvement of 2.6 AP (DUO) on YOLOv5-S~\cite{36}. Meanwhile, UnitModule also significantly improves their performance on the new test dataset \(\text{URPC}_{test}\), achieving higher improvement than on the DUO test set. The results indicate that UnitModule significantly improves the generalization of the object detector in underwater images, enabling it to achieve good performance on brand-new image data.

\textbf{Performance on other object detectors.}
As with YOLO-like object detectors, we also compare the performance of other object detectors with UnitModule including single-stage detectors, two-stage detectors, and DERT, as shown in Table~\ref{table:obr50} and Table~\ref{table:obr101}. Their performance has also been improved by UnitModule, and the performance improvement on \(\text{URPC}_{test}\) is higher than that on DUO.

\begin{figure*}[t]
	\centering
		\includegraphics[width=\linewidth]{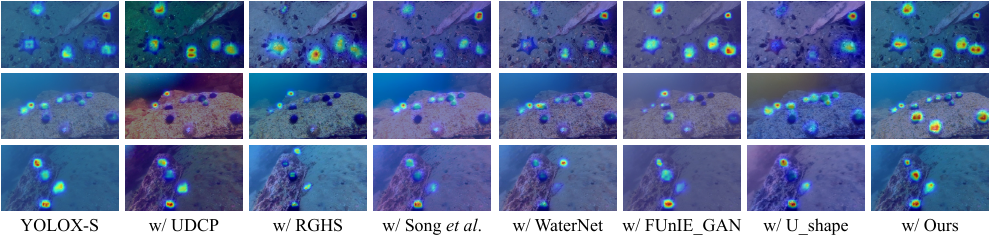}
	\caption{
    Comparison of attention maps in YOLOX-S~\cite{38} trained with different underwater image enhancement methods. The attention maps in the neck of the detector are visualized on the enhanced input image by using the Grad-CAM method~\cite{62}.
    }
	\label{fig:grad cam}
\end{figure*}

Our UnitModule improves the performance of different kinds of detectors in underwater images through joint training. Moreover, as the number of parameters in the model increases, the performance improvement decreases. In Table~\ref{table:obyolo}, models with T, S, M, and L parameter scales exhibit performance improvement from high to low. And models with ResNet-101~\cite{64} backbone have lower performance improvement than that with ResNet-50 in Table~\ref{table:obr50} and Table~\ref{table:obr101}. It suggests that models with a small number of parameters are more likely to gain performance improvement from UnitModule, which reduces the underwater noise in the input image so that the object detection model learns less about the knowledge of generalization noise, allowing the model to focus more on learning object detection. While the models with a large number of parameters themselves have a higher generalization ability, the performance improvement they gain from UnitModule decreases. It also shows that it is advantageous to use UnitModule in models with a small number of parameters and its performance to get higher benefits at a low cost.

In addition, we conduct experiments on MS COCO~\cite{79} dataset to compare the performance of UnitModule in~\ref{appendix:results on ms coco}. The results show that the same object detection model with or without UnitModule has similar performance (only minor fluctuations).  It is demonstrated that our UnitModule can also be used on nearly noiseless image data, but there is no performance improvement, indicating that UnitModule has strong robustness.

\subsection{Quantitative and Visual Analysis}
\label{sec:Quantitative and Visual Analysis}

\begin{table}[t]
\centering

\begin{tabular}{@{}lcc@{}}
\toprule
Method & UCIQE\(\uparrow\) & UIQM\(\uparrow\) \\ 
\midrule

Input                           & 0.501                  & 1.870                  \\ \midrule
UDCP~\cite{71}                  & \textbf{1.611(+1.110)} & 2.590                  \\
RGHS~\cite{72}                  & 1.030                  & 2.834                  \\
Song \textit{et al.}~\cite{50}  & 0.599                  & 2.685                  \\
WaterNet~\cite{68}              & 0.669                  & \textbf{2.886(+1.016)} \\
\(\text{FUnIE\_GAN}\)~\cite{75} & 0.557                  & 1.988                  \\
\(\text{U\_shape}\)~\cite{76}   & 0.969                  & 2.400                  \\
UnitModule (Ours)               & \textbf{0.687(+0.186)} & \textbf{2.604(+0.734)} \\

\bottomrule
\end{tabular}
    \caption{
    Comparison of underwater image quality between different underwater image enhancement methods on DUO~\cite{25}. Higher UCIQE~\cite{69} and UIQM~\cite{70} are better.
    }
    \label{table:image quality}
\end{table}

As shown in Table~\ref{table:image quality}, we use the underwater color image quality evaluation metric (UCIQE~\cite{69}) and the underwater image quality measure (UIQM~\cite{70}) to measure different image enhancement methods. All methods improve the image quality, but the method with the highest image quality does not improve the performance of the detector corresponding to Table~\ref{table:comparison of image enhancement}. Instead, UnitModule which is jointly training with the detector improves both the image quality and the performance of the detector. It indicates that the image preferred by the detector is not the one with the highest image quality. The attention maps in Figure~\ref{fig:grad cam} show that our UnitModule makes the attention area more complete and the attention response to objects higher than others that are not trained jointly with the detector.

\begin{figure}[t]
	\centering
		\includegraphics[width=\linewidth]{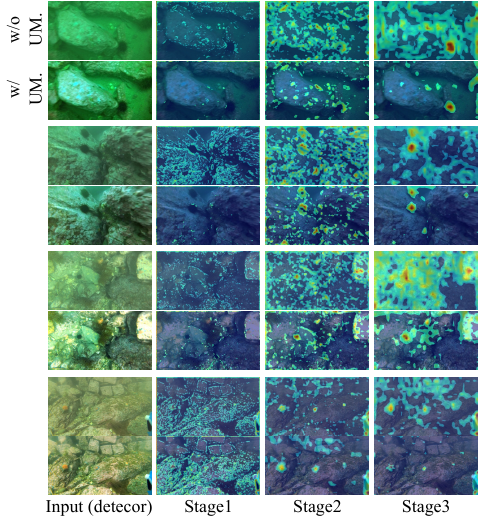}
	\caption{
    Comparison of input images and feature maps in YOLOX-S~\cite{38} between w/o UM. and w/ UM. The input image of the detector in the w/ UM. line is enhanced by UnitModule. The outputs of stages 1-3 in the backbone of the detector are reshaped to feature maps filtered with a fixed threshold for visualization.
    }
	\label{fig:feature map}
\end{figure}
\begin{figure*}[!t]
	\centering
		\includegraphics[width=\linewidth]{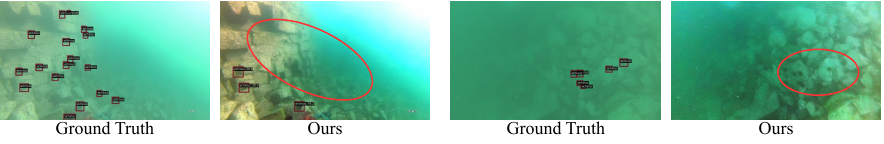}
	\caption{
    Failure cases on DUO~\cite{25}. The red ellipse denotes the missed detections.
    }
	\label{fig:limitation_imgs}
\end{figure*}
\begin{figure}[t]
	\centering
		\includegraphics[width=\linewidth]{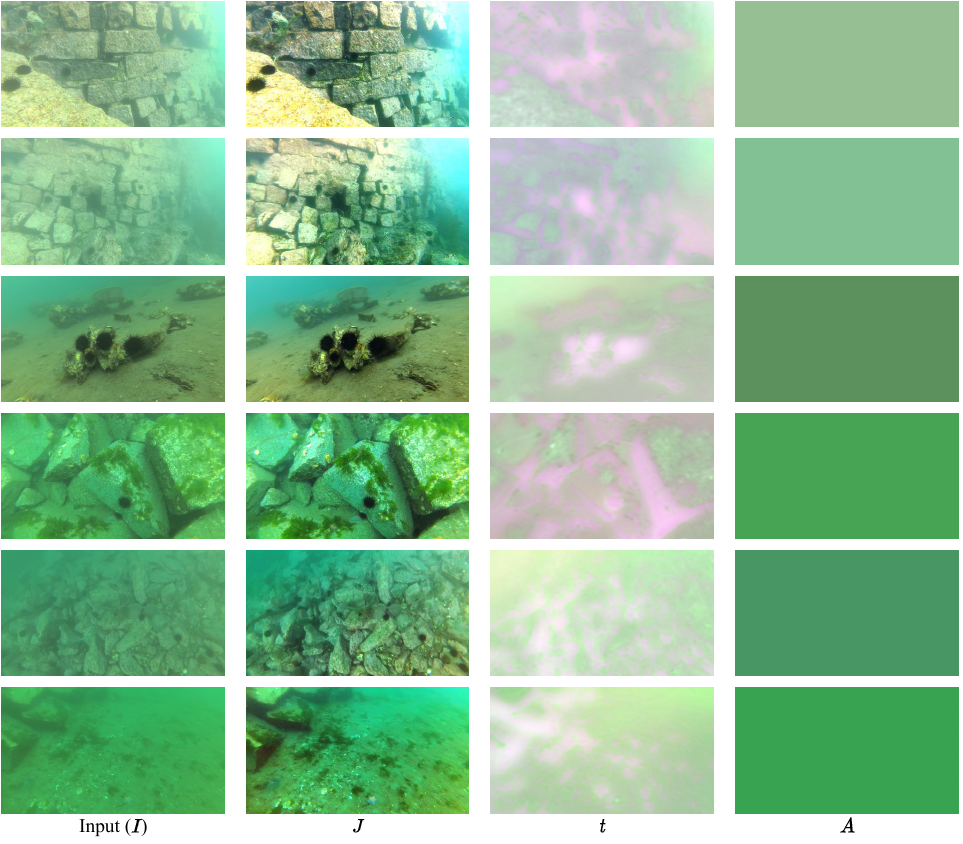}
	\caption{
    The visualization of the input image \(\bm{I}\) for UnitModule with the enhanced image \(\bm{J}\) and its predicted transmission map \(\bm{t}\) and global background light \(\bm{A}\).
    }
	\label{fig:IJtA}
\end{figure}

We visualize the feature maps in the detector trained with and without UnitModule, and the image enhanced by UnitModule in Figure~\ref{fig:feature map}. UnitModule reduces the noise in the input image and makes the enhanced input image become cleaner than before. The enhanced input image reduces the redundant response to the background in the feature map and increases the response to the object. Thanks to the clean image features provided by UnitModule, the detector gets clean features, which reduces the attention of the detector to learning the generalization of noise.

We also visualize the enhanced image \(\bm{J}\), the transmission map \(\bm{t}\), and the global background light \(\bm{A}\) corresponding to the input image \(\bm{I}\) in Figure~\ref{fig:IJtA}. As shown in the figure, we find the transmission map appears more green and blue-purple colors, indicating that the transmittance of green and blue light is higher and the attenuation rate is lower underwater. The predicted transmission map conforms to the laws of natural physics mentioned in Section~\ref{sec:Introduction}, it plays an important role in image enhancement to restore the original color features of the object.

\section{Conclusion}
\label{sec:Conclusion}

In this paper, we propose a plug-and-play lightweight Underwater joint image enhancement Module (UnitModule) to address the problems in underwater object detection using noise reduction methods. Our UnitModule with a small parameter size of 31K causes little delay in inference speed. In particular, the unsupervised learning loss we design allows the detector with UnitModule to perform joint training using only the object detection dataset. Furthermore, a color cast predictor with the assisting color cast loss and a UCRT data augmentation improve the generalization performance of UnitModule on underwater images with different color casts. UnitModule shows significant performance improvement for multiple detectors on the DUO dataset, especially for the detector with a small number of parameters. Besides, UnitModule significantly improves the generalization performance of the detector on new data (as the results on \(\text{URPC}_{test}\)). UnitModule is available for unmanned underwater vehicles for visual tasks.

However, UnitModule still has some limitations. When there are other kinds of noise (such as strong illumination) in the images or strong underwater degradation noise, the proposed method may lead to missed detections, as in Figure~\ref{fig:limitation_imgs}. Our UnitModule is primarily designed to handle degraded underwater noise. Due to the representation of the physical model, its ability to handle other kinds of noise is insufficient, and the lightweight module inevitably results in limited generalization capability for severe underwater degradation noise.

In future work, we can focus on improving the noise reduction capability of the module for other kinds of noise, and strive to build a unified underwater noise reduction module to comprehensively suppress the impact of underwater noise. Our UnitModule can also be easily expanded to other visual tasks, such as underwater semantic segmentation, \textit{etc}. And since UnitModule is based on the atmospheric scattering model, it can also be expanded to similar noise scenes, such as object detection in foggy weather conditions.
\medskip
\newline
\noindent
\textbf{Acknowledgments.}
This research is funded by the National Natural Science Foundation of China, grant number 52371350, by the National Key Laboratory Foundation of Autonomous Marine Vehicle Technology, grant number 2024-HYHXQ-WDZC03, and by the Natural Science Foundation of Hainan Province, grant number 2021JJLH0002.

\appendix

\begin{table*}[!t]
\centering
\resizebox{0.8\linewidth}{!}{

\begin{tabular}{@{}lcc@{}}
\toprule
  Training Config &
  \begin{tabular}[c]{@{}c@{}}YOLOv5, YOLOv6, YOLOv7, \\ YOLOv8, YOLOX, RTMDet\end{tabular} &
  \begin{tabular}[c]{@{}c@{}}Faster R-CNN, Cascade R-CNN, FCOS, \\ RetinaNet, TOOD, DETR, DINO\end{tabular} \\
\midrule

dataset                  & DUO~\cite{25} / MS COCO~\cite{79}           & DUO / MS COCO                                     \\
optimizer                & SGD (AdamW, RTMDet)                         & SGD (AdamW, DETR,DINO)                            \\
base learning rate       & 0.01 (0.004, RTMDet)                        & 0.01 (0.02, Cascade.,Faster.)                     \\
                         &                                             & (0.0001, DETR,DINO)                               \\
weight decay             & 0.0005 (0.05, RTMDet)                       & 0.0001                                            \\
optimizer momentum       & 0.9                                         & 0.9                                               \\
batch size (per GPU)     & 8 (4)                                       & 8 (4)                                             \\
training epochs          & 100                                         & 12 (500, DETR)                                    \\
learning rate schedule   & CosineAnnealingLR                           & MultiStepLR                                       \\
                         & (LinearLR, YOLOv5,YOLOv8)                   &                                                   \\
learning rate decay step & -                                           & [8, 11] ([334], DETR) ([11], DINO)                \\
warmup iterations        & 1000                                        & 500 (0, DETR,DINO)                                \\
input size               & 640 \(\times\) 640                          & 1344 \(\times\) 768                               \\
data augmentation        & RandomFlip                                  & RandomFlip                                        \\
EMA decay                & 0.9999                                      & -                                                 \\
GPU                      & 2 \(\times\) A6000                          & 2 \(\times\) A6000                                \\

\bottomrule
\end{tabular}}
    \caption{
    Training configuration of different object detection models.
    }
    \label{table:training configs}
\end{table*}
\begin{table}[!t]
\centering

\begin{tabular}{@{}lccccc@{}}
\toprule
Method & \(w_{1}\) & \(w_{2}\) & \(w_{3}\) & \(w_{4}\) & \(w_{5}\) \\ 
\midrule

YOLOv5~\cite{36}       & 500  & 0.01 & 0.01 & 0.1 & 0.1 \\
YOLOv6~\cite{37}       & 500  & 0.01 & 0.01 & 0.1 & 0.1 \\
YOLOv7~\cite{81}       & 500  & 0.01 & 0.01 & 0.1 & 0.1 \\
YOLOv8~\cite{82}       & 500  & 0.01 & 0.01 & 0.1 & 0.1 \\
YOLOX~\cite{38}        & 500  & 0.01 & 0.01 & 0.1 & 0.1 \\
RTMDet~\cite{83}       & 500  & 0.01 & 0.01 & 0.1 & 0.1 \\
Faster-RCNN~\cite{28}  & 500  & 0.1  & 0.01 & 0.1 & 0.1 \\
Cascade-RCNN~\cite{63} & 500  & 0.01 & 0.01 & 0.1 & 0.1 \\
FCOS~\cite{41}         & 500  & 0.1  & 0.01 & 0.1 & 0.1 \\
RetinaNet~\cite{40}    & 500  & 0.1  & 0.01 & 0.1 & 0.1 \\
TOOD~\cite{42}         & 500  & 0.1  & 0.01 & 0.1 & 0.1 \\
DETR~\cite{43}         & 1000 & 0.01 & 0.01 & 0.1 & 0.1 \\
DINO~\cite{84}         & 1000 & 0.01 & 0.01 & 0.1 & 0.1 \\

\bottomrule
\end{tabular}
    \caption{
    The loss weights of UnitModule for different object detection models.
    }
    \label{table:loss weights}
\end{table}

\section{Appendix}
\label{appendix:appendix A}

\subsection{UnitModule Loss Weights}
\label{appendix:loss weights}

We set magnitude-balanced UnitModule loss weights for each detector based on the loss values of different detectors in Table~\ref{table:loss weights}.

\subsection{Training Configuration}
\label{appendix:training configuration}

We demonstrate the configuration parameters of each object detection model during training, including its hyper-parameters and some tricks in Table~\ref{table:training configs}.

\subsection{Results on MS COCO}
\label{appendix:results on ms coco}

\begin{table}[!t]
\centering
\resizebox{\linewidth}{!}{

\begin{tabular}{@{}lccc@{}}
\toprule
Method & AP\(\uparrow\) & \(\text{AP}_{50}\)\(\uparrow\) & \(\text{AP}_{75}\)\(\uparrow\) \\ 
\midrule

YOLOv5-S~\cite{36}      & 28.9                & 46.5          & 30.9          \\
w/ UnitModule           & \textbf{29.0(+0.1)} & 46.5          & \textbf{31.2} \\ \midrule
YOLOv6-S~\cite{37}      & 35.1                & 50.2          & 37.9          \\
w/ UnitModule           & \textbf{35.3(+0.2)} & \textbf{50.3} & \textbf{38.1} \\ \midrule
YOLOv7-T~\cite{81}      & 29.8                & 46.3          & 32.0          \\
w/ UnitModule           & 29.8                & \textbf{46.5} & 32.0          \\ \midrule
YOLOv8-S~\cite{82}      & 36.6                & 51.9          & 39.5          \\
w/ UnitModule           & \textbf{36.7(+0.1)} & \textbf{52.3} & 39.5          \\ \midrule
YOLOX-S~\cite{38}       & 34.8                & \textbf{52.7} & 37.3          \\
w/ UnitModule           & 34.8                & 52.6          & \textbf{37.4} \\ \midrule
RTMDet-S~\cite{83}      & \textbf{18.2}       & \textbf{28.5} & 19.1          \\
w/ UnitModule           & 18.0(-0.2)          & 28.2          & \textbf{19.2} \\ \midrule
Faster R-CNN~\cite{28}  & 36.7                & \textbf{57.1} & 39.7          \\
w/ UnitModule           & 36.7                & 56.8          & \textbf{40.0} \\ \midrule
Cascade R-CNN~\cite{63} & 39.7                & 57.7          & 43.4          \\
w/ UnitModule           & \textbf{40.0(+0.3)} & \textbf{57.9} & \textbf{43.7} \\ \midrule
FCOS~\cite{41}          & \textbf{36.7}       & \textbf{55.7} & \textbf{39.1} \\
w/ UnitModule           & 36.6(-0.1)          & 55.5          & 38.9          \\ \midrule
RetinaNet~\cite{40}     & 36.1                & 54.8          & \textbf{38.4} \\
w/ UnitModule           & 36.1                & \textbf{55.1} & 38.3          \\ \midrule
TOOD~\cite{42}          & \textbf{41.5}       & \textbf{58.4} & 45.1          \\
w/ UnitModule           & 41.4(-0.1)          & 58.1          & \textbf{45.4} \\ \midrule
DINO~\cite{84}          & 48.2                & 66.2          & \textbf{52.2} \\
w/ UnitModule           & 48.2                & \textbf{66.4} & 52.1          \\

\bottomrule
\end{tabular}}
    \caption{
    Comparison of different object detection models with UnitModule on MS COCO~\cite{79}. The backbone of the non-YOLO-like detector is ResNet-50~\cite{64} pretrained on ImageNet~\cite{65}.
    }
    \label{table:unitmodule on ms coco}
\end{table}
\begin{figure*}[!t]
	\centering
		\includegraphics[width=\linewidth]{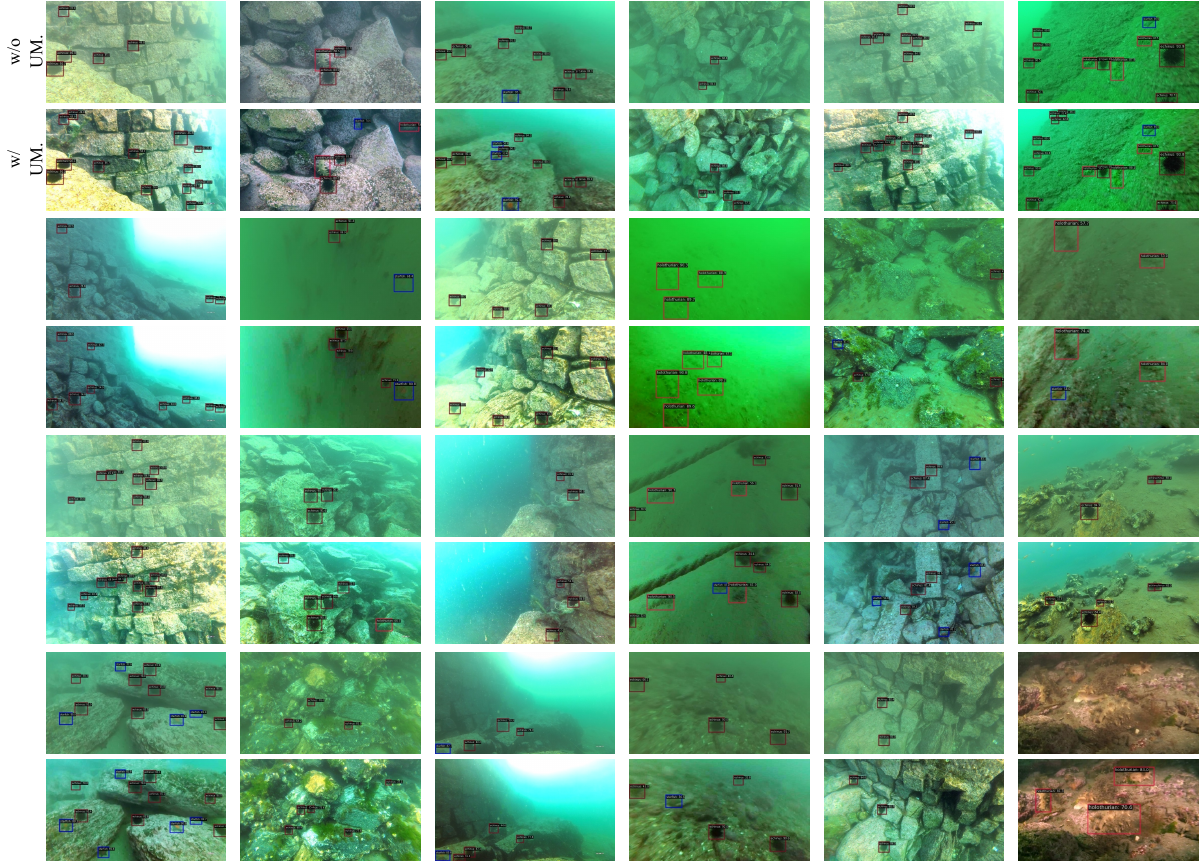}
	\caption{
    Comparison of detection results with and without UnitModule.
    }
	\label{fig:detection}
\end{figure*}

We conduct additional experiments on MS COCO~\cite{79} dataset, which contains about 118K images in the \texttt{train2017} set and 5K images in the \texttt{val2017} set. We compare the performance of different object detection models with or without UnitModule in Table~\ref{table:unitmodule on ms coco}. The results are reported on the \texttt{val2017} set, and UCRT data augmentation is not used when training these object detection models with our UnitModule on the MS COCO.

\subsection{Detection results}
\label{appendix:detection results}

We visualize the detection results of YOLOX-S~\cite{38} without and with UnitModule. The results in Figure~\ref{fig:detection} show that UnitModule improves the detection capability of the detector for underwater objects.

{\small
\bibliographystyle{ieee_fullname}
\bibliography{sections/main}
}

\end{document}